\documentclass{article}
\usepackage{ijcai19}

\usepackage{times}
\usepackage{soul}
\usepackage{url}
\usepackage{multirow}
\usepackage[hidelinks]{hyperref}
\usepackage[utf8]{inputenc}
\usepackage[small]{caption}
\usepackage{graphicx}
\usepackage{amsmath}
\usepackage{amssymb}
\usepackage{bbm}
\usepackage{multirow}
\usepackage{booktabs}
\usepackage{algorithm}
\usepackage{algorithmic}
\usepackage{tikz}
\def\eg {\emph{e.g}.} 
\def\ie {\emph{i.e}.}

\title{COP: Customized Deep Model Compression via Regularized Correlation-Based Filter-Level Pruning}

\author{
Wenxiao Wang$^{1,2}$\and
Cong Fu$^{1,3}$\and
Jishun Guo$^4$\and
Deng Cai$^{1,2}$\footnote{Corresponding author}\And
Xiaofei He$^{1,2}$
\affiliations
$^1$State Key Lab of CAD\&CG, Zhejiang University, Hangzhou, China\\
$^2$Fabu Inc., Hangzhou, China\\
$^3$Alibaba-Zhejiang University Joint Institute of Frontier Technologies, Hangzhou, China\\
$^4$GAC R\&D Center, Guangzhou, China\\
\emails
wenxiaowang@zju.edu.cn,
fc731097343@gmail.com,
guojishun@gacrnd.com,
\{dengcai, xiaofeihe\}@cad.zju.edu.cn
}

\begin{document}

\maketitle

\begin{abstract}

Neural network compression empowers the effective yet unwieldy deep convolutional neural networks (CNN) to be deployed in resource-constrained scenarios. Most state-of-the-art approaches prune the model in filter-level according to the ``importance'' of filters. Despite their success, we notice they suffer from at least two of the following problems: 1) The redundancy among filters is not considered because the importance is evaluated independently. 2) Cross-layer filter comparison is unachievable since the importance is defined locally within each layer. Consequently, we must manually specify layer-wise pruning ratios. 3) They are prone to generate sub-optimal solutions because they neglect the inequality between reducing parameters and reducing computational cost. Reducing the same number of parameters in different positions in the network may reduce different computational cost. To address the above problems, we develop a novel algorithm named as COP (correlation-based pruning), which can detect the redundant filters efficiently. We enable the cross-layer filter comparison through global normalization. We add parameter-quantity and computational-cost regularization terms to the importance, which enables the users to customize the compression according to their preference (smaller or faster). Extensive experiments have shown COP outperforms the others significantly. The code is released at https://github.com/ZJULearning/COP.

\end{abstract}

\section{Introduction} \label{sec:introduction}

The growing demands of deploying deep models to resource-constrained devices such as mobile phones and FPGA have posed great challenges for us. Network pruning has become one of the most popular methods to compress the model without much loss in performance, and methods of network pruning could be divided into two categories: weight-level pruning \cite{han2015deep:scheme,guo2016dynamic,dong2017learning} and filter-level pruning \cite{li2016pruning,liu2017learning,he2018pruning}. 

The weight-level pruning methods try to find out unimportant weights and set them to zeros. In other words, weight-level pruning can compress deep models because it induces sparsity in filters. However, it contributes little to accelerating them unless specialized libraries (such as cuSPARSE) are used. Unfortunately, the support for these libraries on mobile devices, especially on FPGA is very limited. Thus, filter-level pruning methods are proposed to address this problem. They locate and remove unimportant filters in convolutional layers or unimportant nodes in fully connected layers. In this way, both space and inference time cost can be saved a lot.

Undoubtedly, the key factor influencing a filter-level pruning method's performance is how it defines the importance of filters. We find the current state-of-the-art methods have at least two of the following problems.

\begin{figure}[tb]
\flushleft
\includegraphics[scale=0.25]{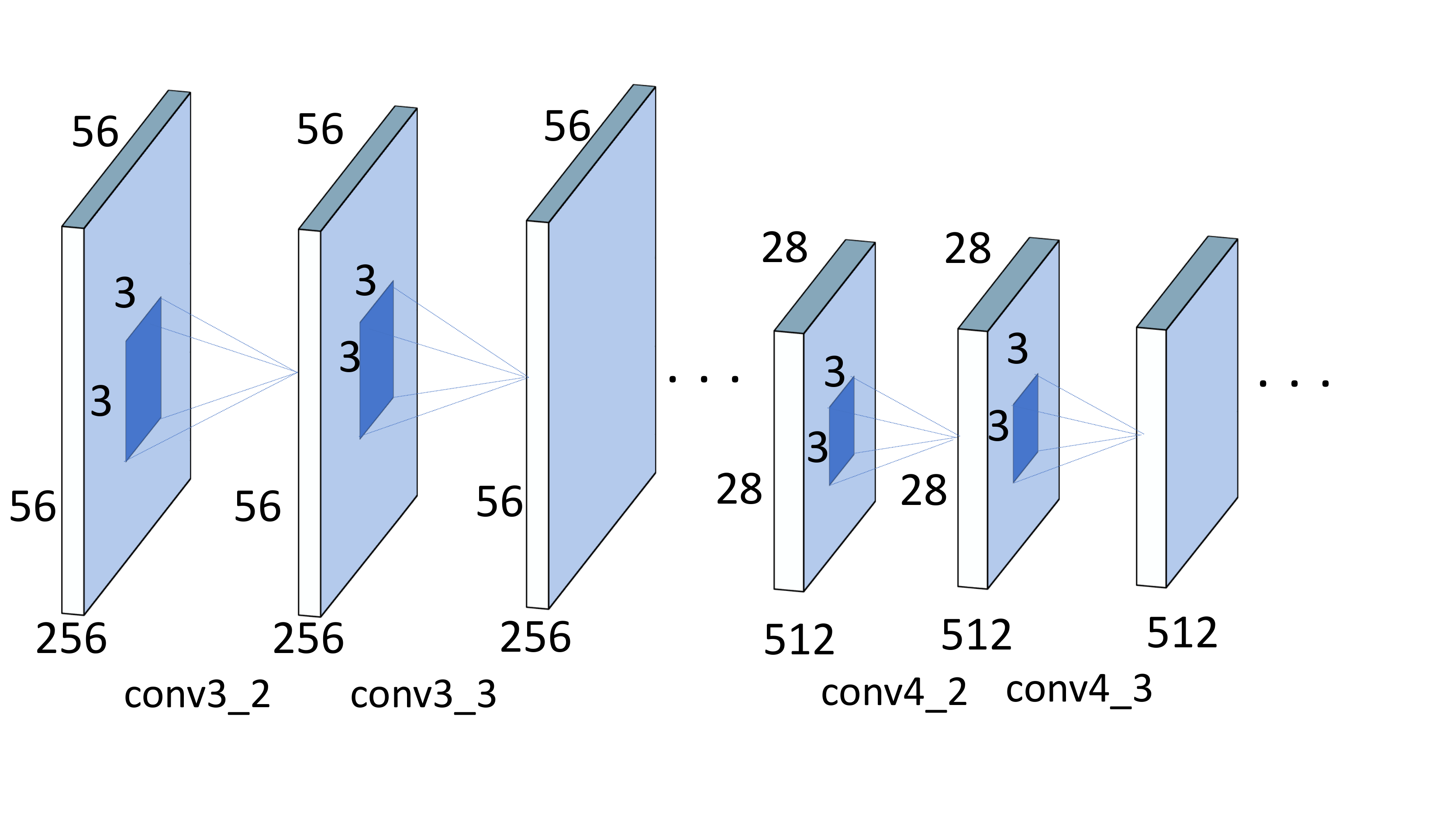}
\caption{The figure is a sketch of VGG16. Consider the following two pruning plans: a) prune 1 filter from conv4\_2; b) prune 2 filters from conv3\_2. With plan a), we can reduce 9216 parameters and 14.5 million FLOPs. With plan b), we can reduce 9216 parameters and 57.8 million FLOPs. a) and b) reduce the same amount of parameters but different computational cost.}
\vspace{-3mm}
\label{fig:vgg}
\end{figure}

\paragraph{High Redundancy.} Deep models with rich parameters may suffer from high redundancy\cite{denil2013predicting} among filters, \ie, different filters may have similar expressibility. Some methods (\eg, \cite{li2016pruning,liu2017learning} only evaluate the importance based on the filter itself and neglect the correlations between filters.  

\paragraph{Lack of Global Vision.} Under the approaches definition of some previous methods (\eg, \cite{li2016pruning,he2018pruning,he2018soft,luo2017thinet:scheme}), the filters can be only compared within the same layer. Thus, it is difficult to specify how many filters should be pruned for each layer.  

\paragraph{Sub-optimal Pruning Strategy.} All the existing methods are prone to generate sub-optimal pruning strategies because they neglect that pruning the same number of parameters may not means reducing the same amount of computation cost. Take Figure \ref{fig:vgg} as an example. Pruning the same number of parameters in different layers of VGG16, we may reduce more computation cost by pruning the lower layers. Previous methods are insensitive to such differences.
\vspace{8pt}

To address the above limitations, we develop a correlation-based pruning algorithm (COP) to detect the redundancy among filters efficiently. We normalize the filter importance of different layers to the same scale to enable global comparison. In addition, we add parameter-quantity and computation-cost regularization terms to enable fine-grained filter pruning. The users can customize the compression simply through weight allocation according to their preference, \ie, whether they want to reduce more parameters or inference complexity. 

It is worthwhile to highlight the advantages of COP:

\begin{itemize}
\item We propose a novel filter-level pruning algorithm for deep model compression, which can a) reduce the redundancy among filters significantly, b) learn proper pruning ratios for different layers automatically, and c) enable fine-grained pruning given users' preference. 
\item Extensive experiments on public datasets have demonstrated COP's advantages over the current state-of-the-art methods. 
\item We also evaluate COP on specially designed compact neural network, MobileNets. COP still produces reasonably good compression ratio with little loss on the inference performance.
\end{itemize}

\section{Related Works}

Network pruning has become one of the most popular methods to compress and accelerate deep CNNs. Network pruning includes two categories: weight-level pruning and filter-level pruning. Weight-level pruning sets unimportant weights in filters to zero, which induces sparsity into filters. However, according to \cite{luo2017thinet:scheme,liu2017learning,molchanov2016pruning}, it is difficult for weight-level pruning to accelerate deep CNNs without specialized libraries or hardware. Filter-level pruning solves this problem, it prunes unimportant filters in convolutional layers, which compresses and accelerates the deep CNNs simultaneously.

In recent years, many filter-level pruning methods have been proposed. ~\cite{li2016pruning} evaluates the importance of filters through the sum of its absolute weights, \ie, $l1$ norm, and decide pruned ratio for each layer manually. ~\cite{luo2017thinet:scheme} evaluates the importance of filters through reconstruction loss and use the same pruned ratio for all layers, \ie, large loss induced by pruning means high importance of the pruned filter. ~\cite{he2017channel} evaluates the importance of filters based on LASSO regression. All the methods mentioned above only evaluate the local importance for filters, \ie, the importance could only be compared within the same layer, so they have to specify the pruned ratio for each layer manually. ~\cite{molchanov2016pruning} evaluates the importance of filters through Taylor expansion, whose value could be compared globally. ~\cite{liu2017learning} also evaluates the global importance of filters based on the scale of batch normalization (BN) layer. However, as we say in Section \ref{sec:introduction}, all these methods neglect the inequality between reducing parameters and reducing computational cost, the users cannot customize the pruned model for different purposes(smaller or faster).

PFA is a filter-level pruning method, which decides the pruned ratio for each layer by performing PCA on feature maps and evaluates the importance of filters by doing correlation analysis on feature maps. PFA is similar to our method because both PFA and COP use correlation to evaluating the importance of filters, however, there are three main differences between PFA and COP: 1) PFA is a data-driven method. It performs the correlation analysis on the activated feature maps, while COP performs correlation analysis on the filter weights. PFA needs all the training data when evaluating the importance of filters, which consumes more computing resources; however, COP only uses the trained model parameters when evaluating filters' importance. 2) PFA is a two-stage pruning method, \ie, it needs to perform PCA first and then correlation analysis when pruning filters; COP only performs correlation analysis. 3) As we mentioned in Section \ref{sec:introduction}, PFA is a sub-optimal pruning method; COP uses two regularization terms to generate fine-grained pruning plans, so users could customize their pruned model for different purposes.

\section{Algorithm} \label{section:algorithm}

\subsection{Symbols and Annotations}

Assuming there are $L$ layers in a deep model, let $P^l$ be the $l_{th}$ layer in the deep model, whose input is $X^l$, output is $Y^l$ and weight is $W^l$. We will omit superscript $l$ for simplicity in the cases of no confusion.

When $P^l$ is a fully connected layer, $W^l$ is of shape $M^l \times N^l$, where $M^l$ is the number of input nodes and $N^l$ is the number of output nodes; $X^l$ is of shape $M^l \times 1$; $Y^l$ is of shape $N^l \times 1$. Let $X^l_{m}$ be the $m_{th}$ node of $X^l$, $Y^l_{n}$ be the $n_{th}$ node of $Y^l$. Let $\vec{\omega}_{m}$ be the $m_{th}$ row of $W^l$, namely $\vec{\omega}_{m} = [W_{m,1},\cdots,W_{m,n}]$.

When $P^l$ is a convolutional layer, $W^l$ is of shape $K^l \times K^l \times M^l \times N^l$, where $K^l$ is the kernel width (assumed to be symmetric), $M^l$ is the number of input channel, $N^l$ is the number of output channel; $X^l$ is of shape $I^l \times I^l \times M^l$; $Y^l$ is of shape $O^l \times O^l \times N^l$; $I^l$, $O^l$ are the size of input and output feature maps respectively. $X^l_{m}$ is the $m_{th}$ input feature map in the $l_{th}$ layer. Let $\vec{\omega}_{i,j,m}$ be the vector $[W_{i,j,m,1},W_{i,j,m,2},\cdots,W_{i,j,m,n}]$. 

$\mu_{\vec{\omega}_{m}}$ is the mean of $\vec{\omega}_{m}$. $\sigma_{\vec{\omega}_{m}}$ is the standard deviation of $\vec{\omega}_{m}$.  Let $E[\cdot]$ be the expectation function.

Note that, in a convolutional layer, pruning feature maps is equivalent to pruning filters, and the importance of feature maps could also be seen as the importance of filters. Therefore, for simplicity, we will not differentiate pruning filters from pruning feature maps.

\subsection{Overview}

We propose three techniques to avoid the three limitations mentioned in Section \ref{sec:introduction}. 1) Evaluate the correlation-based importance of filters to remove redundant filters. 2) Normalize the correlations from all layers to the same scale for global comparison. 3) Adding parameter-quantity and computational-cost regularization terms to the importance evaluation. The users could choose freely whether they want to reduce more parameters or more computational cost by adjusting the weights of two regularization terms.

\begin{figure}[tb]
\flushleft
\includegraphics[scale=0.25]{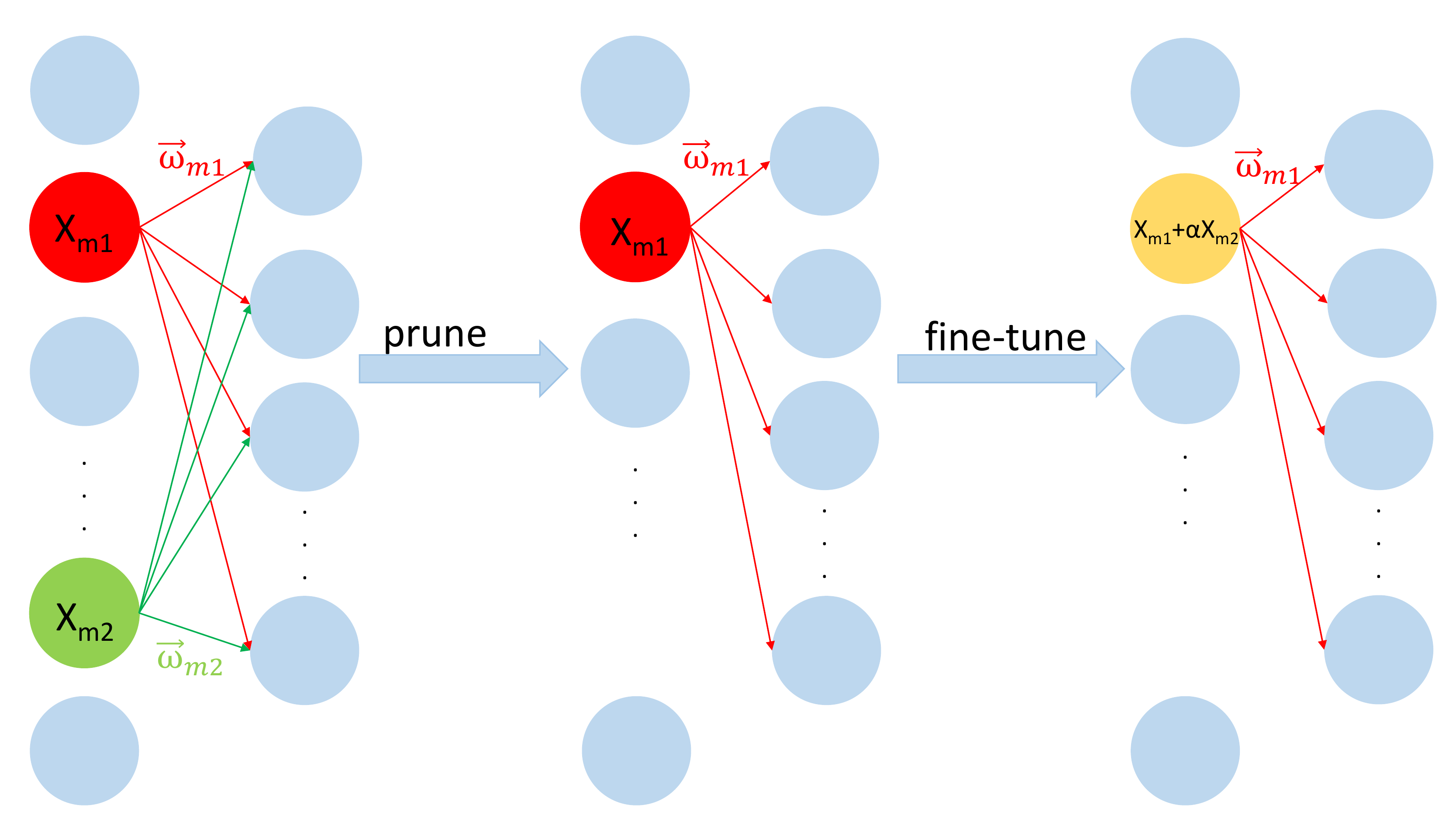}
\caption{The figure is an illustration of the pruning process on a fully connected layer. $X_{m1}$ and $X_{m2}$ are input nodes, $\vec{\omega}_{m1}$ and $\vec{\omega}_{m2}$ are the weights connected with them. If $\vec{\omega}_{m2} \approx \alpha \vec{\omega}_{m1}$, then $X_{m2}$ could be pruned. In the fine-tuned model, $X_{m1}$ would change to $(X_{m1} + \alpha X_{m2})$ because $\vec{\omega}_{m1}X_{m1} + \vec{\omega}_{m2}X_{m2} \approx \vec{\omega}_{m1}(X_{m1}+\alpha X_{m2})$}
\vspace{-3mm}
\label{fig:fully}
\end{figure}

\subsection{Method}
\subsubsection{Normalized Correlation-Based Importance}

We will first take fully connected layers as examples and introduce the calculation of correlation-based importance in detail, then generalize it to convolutional layers.

Figure \ref{fig:fully} is an illustration of our idea. In a fully connected layer, $Y_n$ is calculated as $Y_n = \sum_{m} W_{m,n} X_{m}$, we omit the bias term for simplicity. If $\exists m1, m2 \in [1,M], \alpha \in \mathbb{R}$ such that $\vec{\omega}_{m2} \approx \alpha\vec{\omega}_{m1}$, then $Y_n$ could also be computed as Equation \ref{equ:fully2} for all $n \in [1, N]$. $\epsilon$ is the loss induced because $\vec{\omega}_{m2}$ and $\vec{\omega}_{m1}$ are not strictly linearly related. So, we could merge $X_{m1}$ and $X_{m2}$ through pruning and fine-tuning. 

\begin{equation}
\label{equ:fully2}
\begin{aligned}
&Y_n = \sum_{m \notin \{m1, m2\}} W_{m,n} X_{m} + W_{m1,n} X_{m1} + W_{m2,n} X_{m2} \\
&= \sum_{m \notin \{m1, m2\}} W_{m,n} X_{m} + W_{m1,n} X_{m1} + \alpha W_{m1,n} X_{m2} + \epsilon \\
&= \sum_{m \notin \{m1, m2\}} W_{m,n} X_{m} + W_{m1,n} (X_{m1} + \alpha X_{m2}) + \epsilon \\
\end{aligned}
\end{equation}

$\vec{\omega}_{m2} \approx \alpha \vec{\omega}_{m1}$ implies that $\vec{\omega}_{m1}$ and $\vec{\omega}_{m2}$ are activated in similar patterns. On the other hand, it also implies $X_{m1}$ and $X_{m2}$ express similar information. We propose to measure the redundancy between $\vec{\omega}_{m1}$ and $\vec{\omega}_{m2}$ by using Pearson correlation, which also indicates the similarity between $X_{m1}$ and $X_{m2}$ (Equation \ref{equ:fully3}).

\begin{equation}
\label{equ:fully3}
\begin{aligned}
sim&(X_{m1}, X_{m2}) = corr(\vec{\omega}_{m1},\vec{\omega}_{m2}) \\
&= \frac{E[(\vec{\omega}_{m1} - \mu_{\vec{\omega}_{m1}})(\vec{\omega}_{m2}-\mu_{\vec{\omega}_{m2}})]}{\sigma_{\vec{\omega}_{m1}}\sigma_{\vec{\omega}_{m2}}}
\end{aligned}
\end{equation}

The redundancy evaluation mentioned above could be generalized to convolutional layer with slight modifications because a fully connected layer could be seen as a convolutional layer whose weights are of shape $1 \times 1 \times M \times N$. For a convolutional layer whose filters are of shape $K \times K \times M \times N$, one filter can be seen as a group of $K\times K$ independent nodes. We regroup the filter tensors into $K\times K$ sets of nodes and calculate the pair-wise node-correlations accordingly. Finally, we calculate the filter-correlation by averaging the node-correlations on the respective $K\times K$ nodes of the given filter (as shown in Equation \ref{equ:conv}).

\begin{equation}
\label{equ:conv}
\begin{aligned}
sim(X_{m1}, X_{m2}) = \frac{1}{K^2} \sum_{i}^K \sum_{j}^K corr(\vec{\omega}_{i,j,m1},\vec{\omega}_{i,j,m2})
\end{aligned}
\end{equation}

For two correlated filters, we have to remove one of them. It is hard to tell which is more important if we only look at their correlation. Instead, if one filter is highly correlated with many other filters, we believe it can be removed because other filters can take over its job. Thus, among the correlation coefficients between the given filter and the others, we select the $k$ highest ones and average them to get the importance of the filter. 

\begin{figure}[tb]
\flushleft
\includegraphics[scale=0.25]{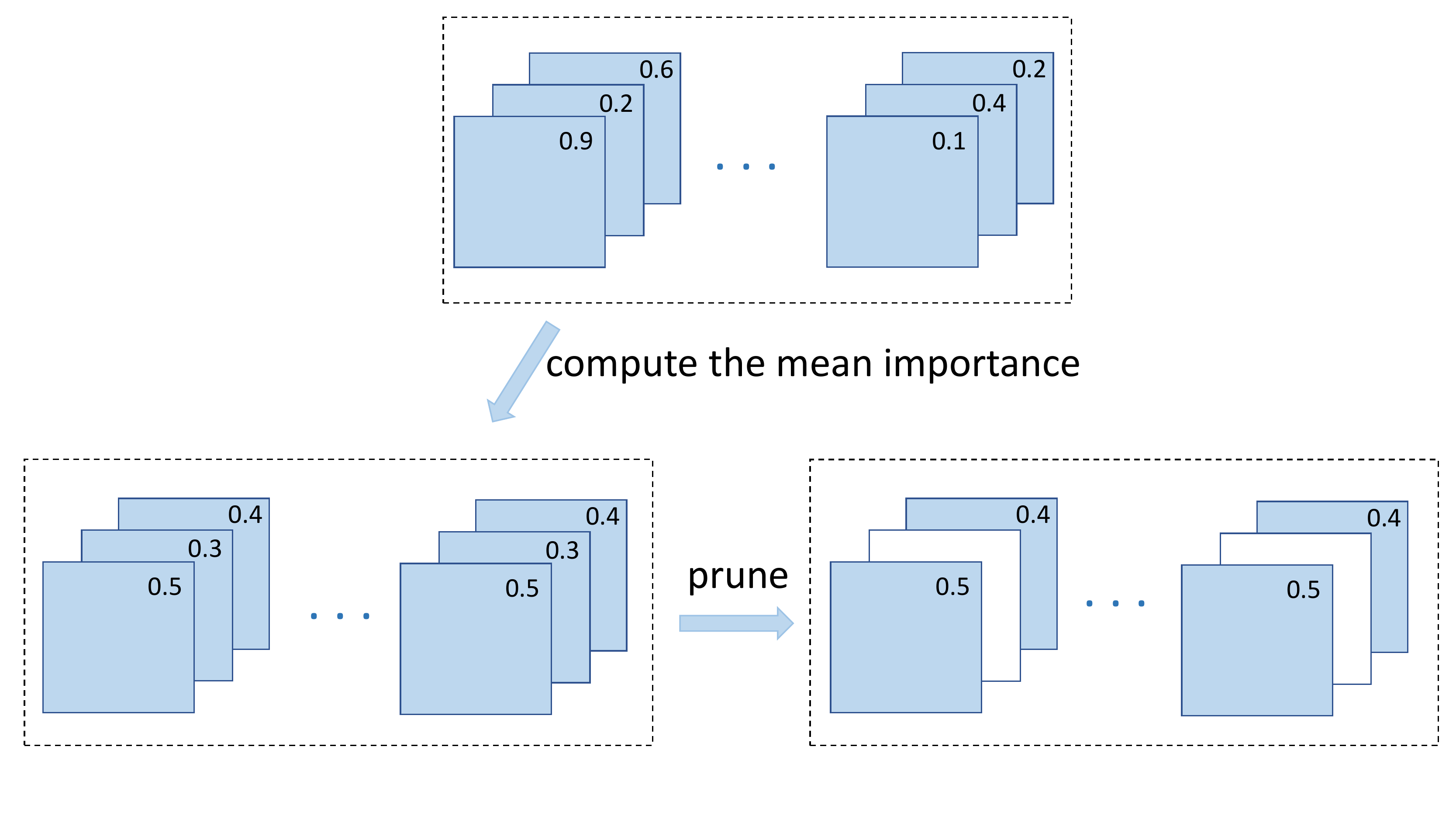}
\caption{The figure is a residual block without linear projection. The input feature maps are on the left of the dashed boxes, and the output feature maps are on the right. The numbers on the feature maps are the importance of feature maps. We compute the mean importance of input feature maps and corresponding output feature maps as overall importance and prune the input and output feature maps simultaneously.}
\vspace{-3mm}
\label{fig:resnet}
\end{figure}

However, the importance defined is still a local one because the correlations are calculated among filters within the same layer. Generally, the correlation distributions of different layers are quite different due to their different functions and scopes. Thus, we need to normalize the importance to enable cross-layer comparison. There are various methods to normalize the importance. \cite{molchanov2016pruning} proposed a simple layer-wise $l2$-normalization which re-scales the importance distributions of different layers. Similarly, one can also normalize different importance distributions with $l1$-normalization. However, we argue that they are not suitable for correlation-based importance because they will cause improper scaling of the importance. For instance, for layers with many filters, the importance of each filter tends to be very small because the denominator is large. Consequently, the procedure will always try to prune the layers with more filters. We propose to use $max$-normalization. Specifically, we normalize the correlation distribution of each layer by dividing them by the maximal importance to align the correlation distributions to $[0,1]$. Formally, we define the normalized importance as Equation \ref{equ:CoI}.

\begin{equation}
\label{equ:CoI}
\begin{split}
Imp(X^l_m) &= 1 - \frac{1}{k} \sum^{k}{TopK(\frac{sim(X^l_{m}, X^l_{n})}{\max_{p\neq q}(sim(X^l_{p}, X^l_{q}))})}, \\
& p,q,n \in [1,M] \, and \, n \neq m\\
\end{split}
\end{equation}

\begin{table}[t]
\centering
\begin{tabular}{c|crrr}
    \toprule
    Dataset & Alg & Acc(\%) & Prr(\%) & Frr(\%)   \\
    \midrule
    
    \multirow{11}{*}{CIFAR10}
    & VGG16 & \textbf{93.56} & 0.0 & 0.0 \\
    & NS$^*$ & \textbf{93.59} & 73.9 & 29.2   \\
    & PFA & 93.16 & 87.3 & 60.9 \\
    & NRE & 93.40 & 92.7 & 67.6 \\
    & SFP$^*$ & 92.99 & 73.0 & 73.0 \\
    & COP & 93.31 & \textbf{92.8} & \textbf{73.5}  \\
    \cline{2-5}
    & ResNet32 & \textbf{92.64} & 0.0 & 0.0 \\
    & NS$^*$ & 90.23 & 35.3 & 53.4   \\
    & SFP & \textbf{92.08} & 39.2 & 41.04 \\
    & PFGM & 91.93 & 52.7 & 53.2 \\
    & COP & 91.97 & \textbf{57.5} & \textbf{53.9}  \\
    \midrule
    
    \multirow{7}{*}{CIFAR100} 
    & VGG16 & \textbf{72.59} & 0.0 & 0.0  \\
    & SFP$^*$ & 71.52 & 42.3 & 42.2 \\
    & PFA & 71.19 & 66.9 & 42.9 \\
    & COP & \textbf{71.77} & \textbf{73.2} & \textbf{43.1}  \\
    \cline{2-5}
    & ResNet32 & \textbf{68.74} & 0.0 & 0.0  \\
    & SFP & 67.83 & 33.8 & 33.9 \\
    & COP & \textbf{68.29} & \textbf{35.2} & \textbf{34.2}  \\
    \midrule
    
    \multirow{7}{*}{ImageNet} 
    & VGG11 & \textbf{63.60} & 0.0 & 0.0 \\
    & NS & \textbf{63.34} & 82.5 & 30.4 \\
    & COP & 62.38 & \textbf{83.2} & \textbf{44.8} \\
    \cline{2-5}
    & ResNet18 & \textbf{70.29} & 0.0 & 0.0 \\
    & SFP & 67.10 & 40.6 & 41.8 \\
    & PFGM & \textbf{67.81} & 40.6 & 41.8  \\
    & COP & 66.98 & \textbf{45.1} & \textbf{43.3} \\
    \bottomrule

\end{tabular}

\caption{The table shows the results of all the algorithms on CIFAR and ImageNet. Alg means the algorithm names. Acc means the classification accuracy, Prr means the parameter-reduction ratio, and Frr means the FLOPs-reduction ratio. Higher is better. VGG and ResNet are the baseline models for pruning. Results with ``*'' are got with their released code, and others are from original papers.}
\vspace{-3mm}
\label{tab:overall-result}
\end{table}

\subsubsection{Regularizers}

As we discussed above, reducing the same number of parameters from different positions in a neural network may reduce quite a different amount of computation cost. To make our approach aware of such differences, we add two regularization terms to enable fine-grained pruning plan generation. As a result, the users can customize the pruning plan through simple weight-allocation. Specifically, we add parameter-quantity and computational-cost regularization terms when evaluating the importance of filters. Pruning a filter of the $l_{th}$ layer influences the parameter quantity and computational cost of the $l_{th}$ layer and the $(l+1)_{th}$ layer, so the parameter quantity ($S^l$) and computational cost ($C^l$) related with the $l_{th}$ layer are defined in Equation \ref{equ:size-cost}. Further, the regularizers are defined in Equation \ref{equ:ReG}. Note that filters in the same layer share the same regularizers, and regularizers for fully connected layers could be computed in a similar way. Adding these two terms to the importance defined above will empower the procedure with the sensitivity of parameter-quantity and computation-cost. Thus, it can generate fine-grained pruning plans.

\begin{equation}
\label{equ:size-cost} 
\begin{aligned}
S^l &= (K^l K^l M^l N^l) + (K^{l+1} K^{l+1} M^{l+1} N^{l+1}) \\
C^l &= 2 I^l I^l K^l K^l M^l N^l \\
&+ 2 I^{l+1} I^{l+1} K^{l+1} K^{l+1} M^{l+1}N^{l+1} \\
\end{aligned}
\end{equation}

\begin{equation}
\label{equ:ReG} 
\begin{aligned}
Reg^l = &\beta(1- \frac {log{(C^l)}}{log(max(C^u))}) \\
+& \gamma(1 - \frac{log({S^l})}{log(max(S^u))}), u \in [1,L]
\end{aligned}
\end{equation}

Finally, we define the regularized importance of $X^l_m$ as Equation \ref{equ:ToI}.

\begin{equation}
\label{equ:ToI} 
\begin{aligned}
ReImp(X_m^l) = Imp(X_m^l) + Reg^l
\end{aligned}
\end{equation}

The COP algorithm is a three-step pipeline :
\textbf{1)} Calculate the regularized importance for all filters and nodes;
\textbf{2)} Specify a global pruning ratio and remove the least important ones according to the ratio;
\textbf{3)} Fine-tune the pruned model with original data. Note that we only prune the network and fine-tune it once. In contrast, many existing methods follow an alternative-prune-and-fine-tune manner.

\begin{table}[t]
\centering
\begin{tabular}{c|crrr}
    \toprule
    Dataset & Alg & Acc(\%) & Prr(\%) & Frr(\%)   \\
    \midrule
    
    \multirow{5}{*}{CIFAR10} 
    & Mob & \textbf{93.90} & 0.0 & 0.0 \\
    & Mob-0.75 & 93.38 & 46.0 & 43.7   \\
    & Mob-0.50 & 92.64 & 74.4 & 74.0   \\
    \cline{2-5}
    & COP-0.50 & \textbf{93.39} & 67.2 & 59.5  \\
    & COP-0.30 & 92.67 & \textbf{83.2} & \textbf{75.6}  \\
    \midrule
    
    \multirow{5}{*}{ImageNet} 
    & Mob & \textbf{66.09} & 0.0 & 0.0 \\
    & Mob-0.75 & 63.44 & 39.0 & 42.8 \\
    & Mob-0.50 & 58.17 & 63.7 & 73.8  \\
    \cline{2-5}
    & COP-0.70 & \textbf{64.52} & 42.9 & 47.0 \\
    & COP-0.40 & 58.39 & \textbf{74.1} & \textbf{79.0} \\
    \bottomrule
\end{tabular}
\caption{The table shows the results of pruning MobileNets on CIFAR10 and ImageNet. Mob means the baseline model. Mob-\textit{x} means thinner MobileNet with width multiplier \textit{x}. COP-\textit{x} means the pruned MobileNets through our approach with filter-preserving ratio $x$.}
\vspace{-3mm}
\label{tab:mobilenet}
\end{table}

\paragraph{Pruning for Depth-Separable Convolutional Layer.} A depth-separable convolutional layer contains a depth-wise convolutional layer and a point-wise convolutional layer. A point-wise convolutional layer is actually a convolutional layer whose filters are of shape $1 \times 1 \times M \times N$, so we prune the point-wise layers as normal convolutional layers. As for depth-wise convolutional layer, the number of input and output feature maps are always the same; the input feature maps of a point-wise layer are also the output feature maps of a depth-wise layer \cite{howard2017mobilenets}. Therefore, pruning the point-wise layer will immediately prune the depth-wise layer; we do not need to prune the depth-wise layer again.

\paragraph{Pruning for Residual Block.} A residual block contains more than one convolutional layer, and the number of input and output feature maps must be equal for a residual block unless there is an extra linear projection in the block \cite{he2016identity}. Thus for all residual blocks which do not contain a linear projection, we compute the mean importance of input and output feature maps as overall importance and prune them simultaneously. Please refer to Figure \ref{fig:resnet} for details.

\section{Experimental Settings} \label{section:experiments}

\begin{table}[t]
\centering
\begin{tabular}{c|lrrr}
\toprule
Network & Algorithm & Acc(\%) & Prr(\%) & Frr(\%) \\
\midrule
\multirow{6}{*}{VGG16}
 & COP$_{cos}$ & 93.11 & 73.0 & 72.6  \\
 & COP$_{dp}$ & 92.27 & 71.9 & 71.1  \\
 & COP$_{cor}$ & \textbf{93.31} & \textbf{92.8} & \textbf{73.5} \\
\cline{2-5}
 & COP$_{l2}$   & 63.60 & 92.2 & 66.5  \\
 & COP$_{l1}$  & 92.98 & 92.5 & 66.7  \\
 & COP$_{max}$  & \textbf{93.31} & \textbf{92.8} & \textbf{73.5} \\
\midrule
\multirow{6}{*}{ResNet32}
 & COP$_{cos}$ & 91.48 & 54.5 & 49.1  \\
 & COP$_{dp}$ & 91.35 & 53.9 & 52.0 \\
 & COP$_{cor}$ & \textbf{91.97} & \textbf{57.5} & \textbf{53.9}  \\
 \cline{2-5}
 & COP$_{l2}$   & 91.51 & 54.0 & 52.2  \\
 & COP$_{l1}$  & 91.43 & 54.7 & 53.5  \\
 & COP$_{max}$ & \textbf{91.97} & \textbf{57.5} & \textbf{53.9}  \\
\bottomrule
\end{tabular}
\caption{The table shows the results of using different redundancy-detectors and global normalization methods in our model. ``$cos$'', ``$dp$'', ``$cor$'' mean cosine, dot-product and correlation redundancy detectors respectively, and ``$l1$'', ``$l2$'' and ``$max$'' are global normalization methods. The experiments are performed on CIFAR10.}
\vspace{-3mm}
\label{tab:ablation}
\end{table}

\subsection{Datasets and Architecture}
We perform the experiments on two well-known public datasets, CIFAR \cite{krizhevsky2009learning} and ImageNet \cite{russakovsky2015imagenet}. Specifically, we use both CIFAR10 and CIFAR100 in the CIFAR collection. 

We test the compression performance of different algorithms on several famous large CNN models, including VGGNet \cite{karen-et-al:scheme}, ResNet \cite{he2016identity}, and MobileNet \cite{howard2017mobilenets}. 

\subsection{Evaluation Protocol}
Following the previous works \cite{jiang2018efficient,liu2017learning,yu2018nisp}, we record the parameter-reduction ratio (Prr) and FLOPs-reduction ratio (Frr) of each algorithm compared with the original model. A higher parameter-reduction ratio means a smaller model size, and a higher FLOPs-reduction ratio means a faster inference. 

\subsection{Compared Algorithms}
We select several recent state-of-the-art methods which are all filter-level pruning algorithms:

\begin{itemize}
    \item \textbf{NS} \cite{liu2017learning} evaluates the importance of filters according to the BN's scaling factors \footnote{https://github.com/Eric-mingjie/network-slimming}.
    
    \item \textbf{NRE} \cite{jiang2018efficient} proposes to prune the model by minimizing the reconstruction error of nonlinear units.
    
    \item \textbf{PFA} \cite{suau2018network} decides the pruned ratio for each layer by performing PCA on feature maps. They evaluate the importance of filters by doing correlation analysis on feature maps.
    
    \item \textbf{SFP} \cite{he2018soft} evaluates the importance of filters with $l2$-norm and prunes filters in a soft manner, i.e., the pruned filters may be retrieved after fine-tuning \footnote{https://github.com/he-y/soft-filter-pruning}.
    
    \item \textbf{PFGM} \cite{he2018pruning} evaluates the importance of filters by analyzing the geometric correlation among the filters within the same layer.
\end{itemize}

We also compare our pruned MobileNets with thinner MobileNets\cite{howard2017mobilenets}. MobileNets are compact deep neural networks which reduce the models' size and computational cost by replacing convolutional layers with depth-separable convolutional layers. \cite{howard2017mobilenets} also proposes thinner MobileNets to balance accuracy and resources consumption.

\begin{table}[t]
\label{dataset}
\centering
\begin{tabular}{crrr}
\toprule
Algorithm & Acc(\%) & Prr(\%) & Frr(\%) \\
\midrule
VGG16  & \textbf{72.59}  & 0.0 & 0.0  \\
COP($\gamma=0$, $\beta=0$)  & 72.09  & 57.9  & 43.2 \\
COP($\gamma=0$, $\beta=3$)  & 72.17  & 46.0  & \textbf{48.8} \\

COP($\gamma=1$, $\beta=1$)  & 71.98  & 63.8 & 47.4  \\
COP($\gamma=3$, $\beta=0$)  & 71.77  & \textbf{73.2} & 43.1  \\
\midrule
ResNet32  & \textbf{68.74}  & 0.0 & 0.0 \\
COP($\gamma=0$, $\beta=0$)  & 68.09  & 33.0 & 32.9  \\
COP($\gamma=0$, $\beta=3$)  & 68.29  & 35.2 & \textbf{34.2}  \\

COP($\gamma=1$, $\beta=1$)  & 68.21  & 40.4 & 22.9  \\
COP($\gamma=3$, $\beta=0$)  & 68.26  & \textbf{41.1} & 18.6  \\
\bottomrule
\end{tabular}
\caption{The table shows the results of using different $\beta$ and $\gamma$ on CIFAR100. $\beta$ is the weight of computational-cost regularization terms, and $\gamma$ is the weight of parameter-quantity regularization terms.}
\vspace{-3mm}
\label{tab:beta-gamma}
\end{table}

\begin{figure*}[tb]
\centering
\includegraphics[scale=0.23]{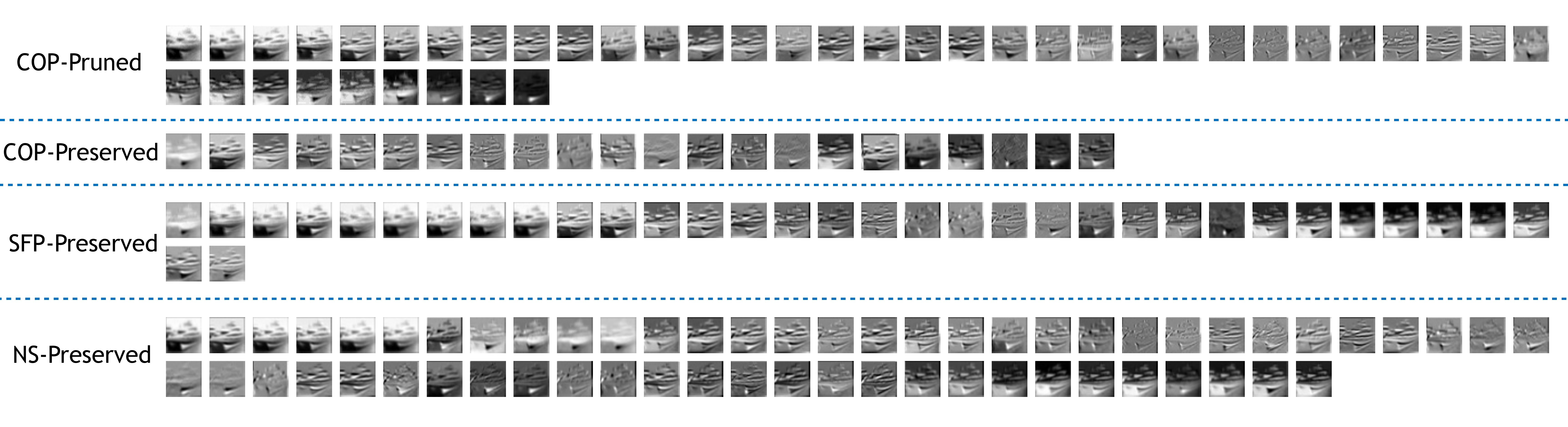}
\caption{The figure shows the output feature maps of ``conv1\_1'' in VGG16; the input is a random image from CIFAR10. COP-pruned means feature maps which are pruned by COP, COP-preserved means feature maps which are not pruned. SFP-preserved and NS-preserved mean feature maps which are preserved by these two algorithms. As the figure shows, COP-pruned contains many feature maps which look similar, but feature maps in COP-preserved are less similar to each other. However, SFP-preserved and NS-preserved still contain many feature maps which are similar, so the model pruned by these two algorithms are still of high redundancy. The figure implies that COP eliminates more redundancy of CNNs than SFP and NS.}
\vspace{-3mm}
\label{fig:visualize}
\end{figure*}

\subsection{Configuration}

We set the batch-size of the stochastic gradient descent algorithm (SGD) to be 128, the momentum coefficient to be 0.9 for all models on CIFAR. For the VGG16 model on CIFAR, we use a weight decay rate of 0.0015, and we use a weight decay rate of 0.0006 and 0.0002 for the ResNet32 and MobileNet respectively on CIFAR.

When training on ImageNet dataset, we use the same architecture and training policy with the original paper~\cite{karen-et-al:scheme} for VGG11. We use the same architecture with the original paper~\cite{howard2017mobilenets} for MobileNets but the same training policy with VGG11 because it takes too much time to train MobileNets with the official training policy.

We use $k=3$ for $TopK$ when pruning the model.

\section{Results and Analysis}
\label{section:results}

\subsection{Compression Efficiency}
For the methods which do not report the results on given datasets and do not release their code either, we pass them in the experiments. The performance of different algorithms on the three datasets is given in Table \ref{tab:overall-result}, we experiment with several pruned ratio on COP and choose the maximal pruned ratio under the constraint of acceptable accuracy loss. On CIFAR, all the algorithms prune the VGG16 and ResNet32 models to get their compressed versions. On ImageNet, all the algorithms prune the VGG11 and ResNet18 models to get their compressed versions. As is shown in Table \ref{tab:overall-result}, we can draw some conclusions as follows:

\begin{enumerate}
    \item COP achieves a higher compression ratio and speedup ratio with similar accuracy to other algorithms. Especially, COP gets the highest numbers on all three metrics on CIFAR100.
    \item COP can prune much more parameters than most of the compared algorithms (\eg, NS, SFP, and PFGM) on different datasets and architecture because COP can detect the redundancy among filters better.
    \item For the algorithms which can prune the similar number of parameters to COP (\eg, NRE and PFGM), COP gets more speed-up, which is the contribution of the computation-cost and parameter-size regularizers.

\end{enumerate}
Though the MobileNet is a specially designed compact network, we can still prune it without much performance loss. The results are given in Table \ref{tab:mobilenet}. MobileNet has its own way to compress itself \cite{howard2017mobilenets}. We can see that COP can get better numbers in all the metrics.

\subsection{Ablation Study}
\paragraph{Redundancy Detector.} We have tried dot-product similarity, cosine similarity, and Pearson correlation as the redundancy detectors, and also tried $l1$-, $l2$-, and $max$-normalization methods as discussed above. The results are shown in Table \ref{tab:ablation}. We can see that Pearson correlation and $max$-normalization are the best configurations because the Pearson correlation detect the redundancy better and the $max$-normalization is insensitive to the layer-width.
\paragraph{Regularizer Efficiency.} We try different weights for two regularization terms, and the results are shown in Table \ref{tab:beta-gamma}. We can see that the regularizer do have significant impacts on the compression. Larger $\gamma$ reduces more parameters while larger $\beta$ reduces more computation cost, which is exactly consistent with our expectation.

\subsection{Case Study}

COP focuses on observing redundant filters, so our pruned models contain less redundancy than others. With less redundancy, feature maps in the model mainly express diversified information. Figure \ref{fig:visualize} is an example of visualized feature maps of ``conv1\_1'' in VGG16. Every feature map pruned by COP expresses similar information to those preserved. We compare COP with SFP and NS. SFP and NS can prune many unimportant feature maps, but there are still some similar feature maps being preserved.

\section{Conclusion} \label{section:conclusion}

We propose a novel filter-level pruning method, COP, to address the limitations of previous works in the following aspects: removing redundancy among filters; enabling cross-layer importance comparison; generating fine-grained pruning strategies (sensitive to desired computation cost and model size). Extensive experiments have shown our significant advantages over other state-of-the-art methods. Moreover, COP can also prune specially designed compact networks such as MobileNet and get larger compression ratio and speedup ratio than its own compression method.

\bibliographystyle{named}
\bibliography{ijcai19}

\end{document}